\documentclass[conference]{IEEEtran}
\IEEEoverridecommandlockouts
\usepackage{cite}
\usepackage{amsmath,amssymb,amsfonts}
\usepackage{algorithmic}
\usepackage{graphicx}
\usepackage{textcomp}
\usepackage{xcolor}
\def\BibTeX{{\rm B\kern-.05em{\sc i\kern-.025em b}\kern-.08em
    T\kern-.1667em\lower.7ex\hbox{E}\kern-.125emX}}
    
\begin{document}

\title{Progressive Co-Attention Network for Fine-Grained Visual Classification}

%\author{\IEEEauthorblockN{Tian Zhang, Dongliang Chang, Zhanyu Ma$^*$, %and Jun Guo}
%\IEEEauthorblockA{Pattern Recognition and Intelligent System Lab.,\\ %Beijing University of Posts and Telecommunications, Beijing, China}
%\thanks{$^*$ Corresponding author}
%}

\author{\IEEEauthorblockN{Tian Zhang$^1$, Dongliang Chang$^1$, Zhanyu Ma$^{1,2,*}$, and Jun Guo$^1$}
\IEEEauthorblockA{$^1$ Pattern Recognition and Intelligent System Lab.,\\ Beijing University of Posts and Telecommunications, Beijing, China}
\IEEEauthorblockA{$^2$ Beijing Academy of Artificial Intelligence, Beijing, China}
\thanks{$^*$ Corresponding author}
}

\maketitle

\begin{abstract}
Fine-grained visual classification aims to recognize images belonging to multiple sub-categories within a same category. It is a challenging task due to the inherently subtle variations among highly-confused categories. Most existing methods only take an individual image as input, which may limit the ability of models to recognize contrastive clues from different images. In this paper, we propose an effective method called progressive co-attention network (PCA-Net) to tackle this problem. Specifically, we calculate the channel-wise similarity by encouraging interaction between the feature channels within same-category image pairs to capture the common discriminative features. Considering that complementary information is also crucial for recognition, we erase the prominent areas enhanced by the channel interaction to force the network to focus on other discriminative regions. The proposed model has achieved competitive results on three fine-grained visual classification benchmark datasets: CUB-$200$-$2011$, Stanford Cars, and FGVC Aircraft. 
\end{abstract}


\section{Introduction}
In the past few years, convolutional neural networks (CNNs)~\cite{krizhevsky2012classification,Simonyan2014recognition,szegedy2015going,he2016deep} have achieved remarkable success in general image classification tasks. However, recognizing fine-grained object categories (e.g., bird species~\cite{wah2011caltech}, car~\cite{krause20133d} and aircraft~\cite{maji13fine} models) is still a challenging task due to high intra-class variances and low inter-class variances, which attracts extensive research attentions. 

\begin{figure}[t]
\centering
\includegraphics[height=4.6cm]{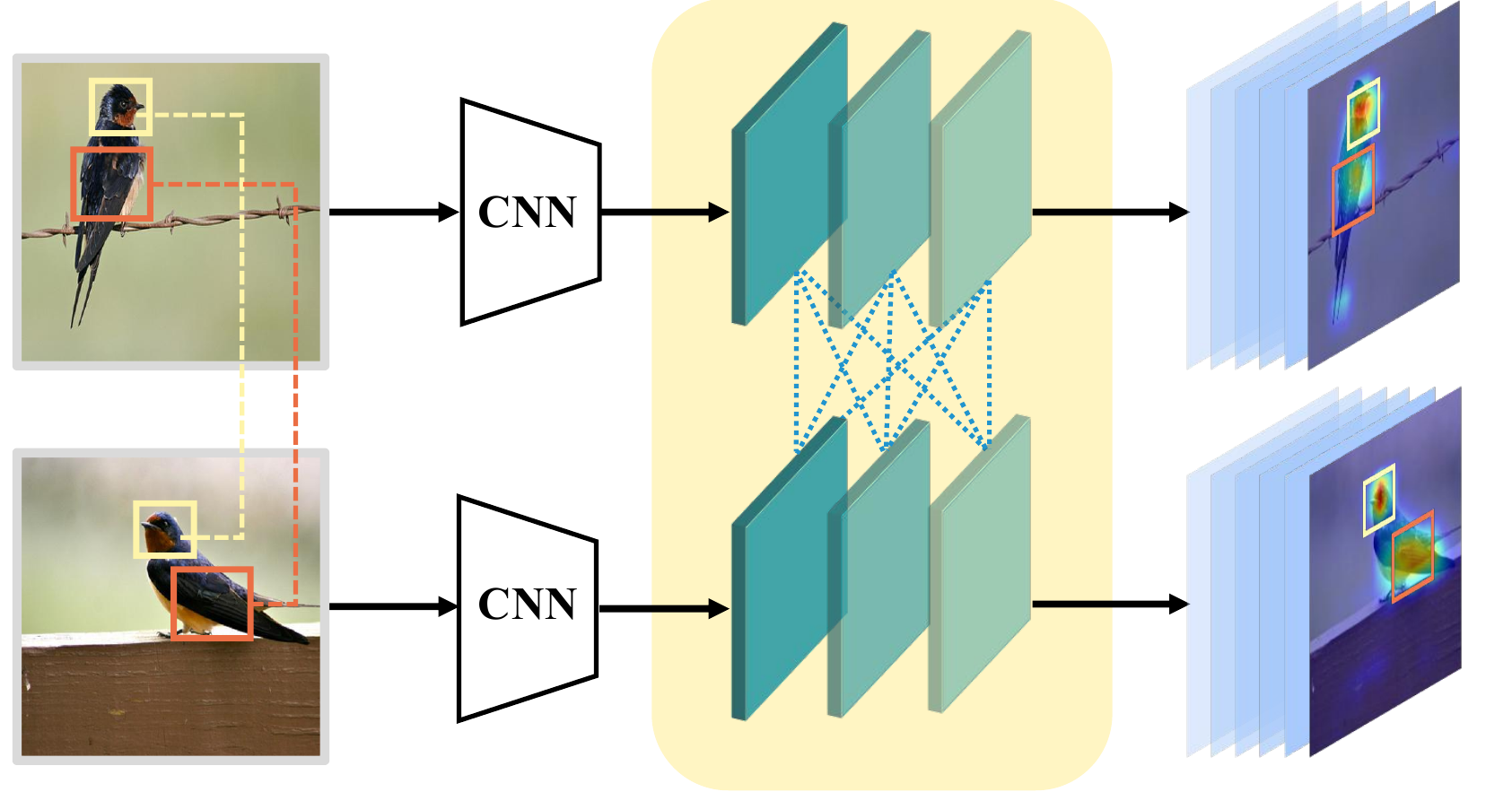}
\caption{The motivation of the proposed method. Most previous methods only took an individual image as input, and the relationship between images was not explored. In our work, we input a pair of same-category images and model the channel interaction between them to capture their common features.
}
\label{fig:schematic}
\end{figure}

The research of fine-grained visual classification has changed from multi-stage framework with hand-crafted features~\cite{berg2013poof,yao2012codebook,chai2013symbiotic,gosselin2014revisiting} to multi-stage framework based on CNN features and then to end-to-end methods. Some works design a localization sub-network to locate key parts, and then a classification sub-network follows for identification. Such as STN~\cite{jaderberg2015spatial}, RA-CNN~\cite{fu2017look}, NTS-Net~\cite{yang2018learning}. They first detected the local areas, and then cropped the detected areas on the original image to learn the key parts. However, little or no effort has been made to guarantee maximum discriminability in these areas~\cite{du2020fine} and the training procedures of these methods are sophisticated due to the complex architecture designs~\cite{sun2018multi}. Others directly learn a more discriminative feature representation by developing powerful deep models~\cite{lin2015bilinear,gao2016compact,kong2017low,cui2017kernel,cai2017higher}. Among them, the most representative method is bilinear CNNs~\cite{lin2015bilinear}, which used two deep convolutional networks to fuse output features, so that it could encode high-order statistics of convolutional activation. More recent advances reduce high feature dimensions~\cite{gao2016compact,kong2017low} or use kernel methods~\cite{cui2017kernel,cai2017higher} to extract higher order information. However, these works do not have an effective design for fine-grained visual classification in terms of the relationship between categories and the number of key parts. Meanwhile, most of the methods mentioned above only take an individual image as input, which will limit their ability to identify contrastive clues from different images for fine-grained visual classification. Generally, humans often recognize fine-grained objects by comparing image pairs, instead of checking single image alone~\cite{zhuang2020learning}. As shown in Figure~\ref{fig:schematic}, by comparing 
a pair of same-category images, we can easily capture their common and prominent features, such as head and wings. In this way, we can learn the discriminative features precisely. 

Inspired by this observation, we propose a co-attention module (CA-Module) to model the channel interaction between a pair of same-category images. By capturing the contrastive features between channels, the model can better learn the commonality of same-category images, thereby force the network to focus on the common discriminative features. However, only focusing on common features of same-category images will cause the network to ignore complementary features that are essential for highly-confused categories. In order to tackle this problem, we propose an attention erase module (AE-Module) to learn complementary features by erasing the most prominent area found in CA-Module. With the combination of these two modules, the proposed method can capture more relevant areas to improve the model performance. 

\begin{figure*}[ht]
\centering
\includegraphics[height=7cm]{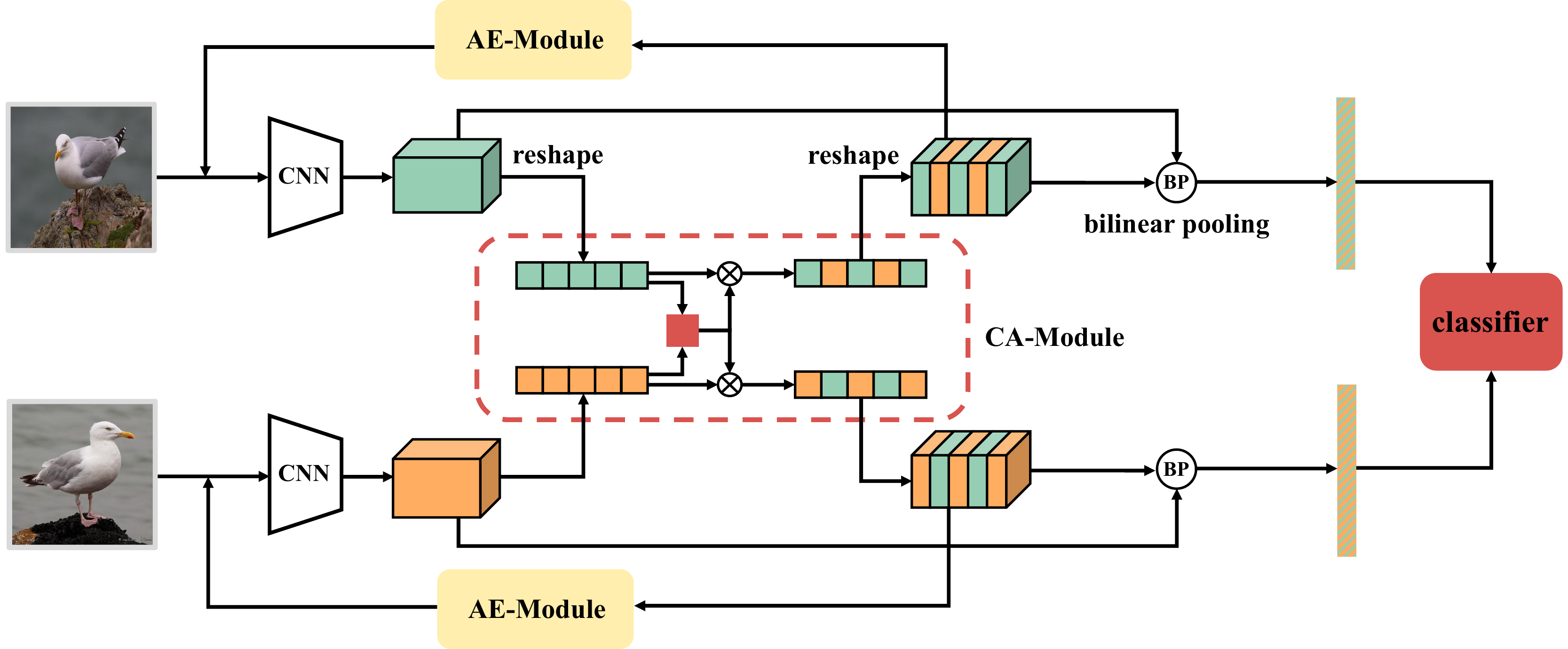}
\caption{The framework of the progressive co-attention network (PCA-Net). The CA-Module can model the channel-wise interaction within a pair of same-category images to focus on the prominent areas with common characteristics. The AE-Module can erase the images to distract attention to other areas to capture complementary features.}
\label{fig:framework}
\end{figure*}

\section{METHODOLOGY}
In this section, we present the progressive co-attention network (PCA-Net) for fine-grained visual classification, as illustrated in Figure~\ref{fig:framework}. It is made up of the CA-Module and the AE-Module to dig distinguishing features that are both prominent and complementary.

\subsection{Co-Attention Module}\label{AA}
Each channel of feature maps can be considered as the representation of a certain feature, and its value is the response to the current feature's strength. Convolutional layer is to model the interaction between channels within an image to generate more abundant features. Gao et al.~\cite{2020CIN} proposed a self-channel interaction (SCI) module to model the interplay between different channels within an image more effectively. Since the intra-class variances of fine-grained images are very high, it is important to compare a pair of same-category images to obtain common features. Hence, unlike~\cite{2020CIN} modeling self-channel interaction, we propose a co-attention module (CA-Module) to model the interaction between channels within two same-category images, forcing the network to focus on the common areas.

Specifically, given an image pair, the two images are first processed by a convolutional network to generate a pair of feature maps ${F_{1}, F_{2}} \in \mathbb{R}^{c\times h\times w}$. The $c$, $h$ and $w$ indicate channel numbers, height, and width respectively. We reshape the feature maps $F_{1}, F_{2}$  to  $F^{'}_{1}, F^{'}_{2}  \in \mathbb{R}^{c \times l},l=h \times w$. Then the channel-wise similarity is calculated by performing a bilinear operation on  $F^{'}_{1} $ and  $ F_{2}^ \mathrm{ 'T } $ to obtain a bilinear matrix $F^{'}_{1} F_{2}^ \mathrm{ 'T } $ as $M$. Then we take the negative value of it and get the weight matrix through a softmax function:
\begin{eqnarray}
W_{ij}=\frac{exp({-M}_{ij})}{\sum_{k=1}^{c} exp({-M}_{ik})}.
\end{eqnarray}
Where $i$ represents the $i^{th}$ channel of the first image, and $j$ represents the $j^{th}$ channel of the second image. We weight the weight matrix $W$ to the original feature maps. Then the interacted feature maps can be obtained as:
\begin{eqnarray}
F_{W1}= {W \times F_{1}}\in \mathbb{R}^{c \times h \times w}, \nonumber \\
F_{W2}= {W \times F_{2}}\in \mathbb{R}^{c \times h \times w}.
\end{eqnarray}

Moreover, we use the cross-entropy loss for classification based on the predictions that are generated by the features $F_{W}$.

After weighting the feature maps, the channels corresponding to similar features are enhanced to find the commonality of this pair of images. The response values of similar parts increase, and the response values of other parts decrease.

\subsection{Attention Erase Module}

The CA-Module aims to focus on the common features of same-category images, but it ignores other slight clues containing complementary information. Because there are subtle differences between highly-confused categories in fine-grained images, it is necessary to pay attention to subtle features. Hence, we propose an attention erase module (AE-Module) to capture the subtle complementary features by erasing the prominent areas in the image.

We perform global average pooling on the feature maps weighted by CA-Module. Then we select the channel of the feature maps corresponding to the maximum value as attention map, and up-sample it to the original image size:
\begin{eqnarray}
A(x,y) = Upsample(\underset{m}{max} \ [F_{W}(m,x,y)]).
\end{eqnarray}
Where $m$ denotes the channel index, $x$ and $y$ denote the spatial indexs.

We obtain a drop mask $M$ by setting the elements $A(x,y)$ larger than threshold $\theta$ to $0$, and setting other elements to $1$:
\begin{eqnarray}
M(x,y) = \begin{cases} 0, & \mbox{if }A(x,y)>\theta \\ 1, & \mbox{else } \end{cases}.
\end{eqnarray}

Overlay the drop mask on the original image to obtain a new image with partial areas erased:
\begin{eqnarray}
I_{e}=I(x,y)\otimes M(x,y),
\end{eqnarray}
where $\otimes$ denotes element-wise multiplication, $I$ denotes the original image, $I_{e}$ denotes the erased image.
\begin{table}
\vspace{-2mm}
\begin{center}
\caption{Statistics of benchmark datasets} \label{tab:datasets}
\begin{tabular}{c|c|c|c}
  \hline
  Datesets & Class & Training & Testing  \\
  \hline
  \hline
  CUB-$200$-$2011$ & $200$ & $5997$ & $5794$ \\
  %\hline
  Stanford Cars & $196$ & $8144$ & $8041$ \\
  %\hline
  FGVC Aircraft & $100$ & $6667$ & $3333$  \\
  \hline
\end{tabular}
\end{center}
\vspace{-3mm}
\end{table}

As the prominent areas of the image are erased, the attention is distracted and the network is forced to learn discriminative information from other areas. It can also reduce the dependence on training samples to improve the robustness of the model.
\begin{table*}[ht]
\begin{center}
\caption{Comparative experiment between recent methods with our method on CUB-$200$-$2011$, Stanford Cars, and FGVC Aircraft. The best result is colored in red, and the second best result is colored in blue.} \label{tab:results}
\begin{tabular}{c|c|c|c|c|c}
  \hline
  Method & Backbone & Input size & CUB-$200$-$2011$ (\%) & Stanford Cars (\%) & FGVC-Aircraft (\%) \\
  \hline
\hline
FT VGG-$19$~\cite{wang2018learning} & VGG-$19$ & $448 \times 448$ & $77.8$ & $84.9$ & $84.8$ \\
RA-CNN~\cite{fu2017look} & VGG-$19$ & $448 \times 448$ & $85.3$ & $92.5$ & $-$ \\
MA-CNN~\cite{zheng2017recognition} & VGG-$19$ & $448 \times 448$ & $86.5$ & $92.8$ & $89.9$ \\
  \hline
  FT ResNet-$50$~\cite{wang2018learning} & ResNet-$50$ & $448 \times 448$ & $84.1$ & $91.7$ & $88.5$ \\
    RAM~\cite{Li_2017_ICCV} & ResNet-$50$ & $448 \times 448$ & $86.0$ & $93.1$ & - \\

      DFL-CNN~\cite{wang2018learning} & ResNet-$50$ & $448 \times 448$ & $87.4$ & $93.1$ & $91.7$ \\

      NTS-Net~\cite{yang2018learning} & ResNet-$50$ & $448 \times 448$ & $87.5$ & $93.9$ & $91.4$ \\
        MC-Loss~\cite{chang2020mc} & ResNet-$50$ & $448 \times 448$ & $87.3$ & $93.7$ & $92.6$ \\

      TASN~\cite{Zheng_2019_CVPR} & ResNet-$50$ & $448 \times 448$ & $87.9$ & $93.8$ & - \\
      Cross-X~\cite{luo2019cross} & ResNet-$50$ & $448 \times 448$ & $87.7$ & $94.6$ & $92.6$ \\
      ACNet~\cite{2019ACNet} & ResNet-$50$ & $448 \times 448$ & $88.1$ & $94.6$ & $92.4$ \\
    \hline
        MAMC~\cite{sun2018multi} & ResNet-$50$ & $448 \times 448$ & $86.2$ & $92.8$ & - \\
    MAMC~\cite{sun2018multi} & ResNet-$101$ & $448 \times 448$ & $86.5$ & $93.0$ & - \\
    \hline
      CIN~\cite{2020CIN} & ResNet-$50$ & $448 \times 448$ & $87.5$ & $94.1$ & $92.6$ \\
        CIN~\cite{2020CIN} & ResNet-$101$ & $448 \times 448$ & $88.1$ & $94.5$ & $92.8$ \\
    \hline
      API-Net~\cite{zhuang2020learning} & ResNet-$50$ & $448 \times 448$ & $87.7$ & \textcolor[RGB]{0,0,255}{\textbf{94.8}} & \textcolor[RGB]{0,0,255}{\textbf{93.0}} \\
      API-Net~\cite{zhuang2020learning} & ResNet-$101$ & $448 \times 448$ & \textcolor[RGB]{0,0,255}{\textbf{88.6}} & $\textcolor[RGB]{255,0,0}{\textbf{94.9}}$ & \textcolor[RGB]{255,0,0}{\textbf{93.4}} \\
  \hline
      Ours & ResNet-$50$ & $448 \times 448$ & $88.3$ & $94.3$ & $92.4$ \\
    Ours & ResNet-$101$ & $448 \times 448$ & \textcolor[RGB]{255,0,0}{\textbf{88.9}} & $94.6$ & $92.8$ \\
  \hline
\end{tabular}
\end{center}
\end{table*}

\subsection{Fine-Grained Feature Learning}
In order to make the learned features more aggregated in embedding space, we introduce the center loss~\cite{wen2016discriminative}. For each category, a feature vector is calculated as the center of the corresponding category, and it will be continuously updated during training phase. By penalizing the deviation between the bilinear feature vector of each sample and the center of the corresponding category, the samples belonging to the same category are grouped together as many as possible, which enhances the discrimination of the learned features.

Benefiting from global and local informative features of an image, we define the feature representations for each image: $F_{J}=\{F_{O},F_{W},F_{E}\}$, where $F_{O}$ denotes the feature maps extracted from the original image, $F_{W}$ denotes the weighted feature maps of the original image, and $F_{E}$ denotes the feature maps extracted from the erased image. These features are then fed to a fully-connection layer with a softmax function for the final classification.

During training phase, the whole model is optimized by losses defined as
\begin{eqnarray}
L=\sum_{i\in I}L_{cls}(Y^{i},Y^{*})+\lambda L_{cen}(Y^{i},Y^{c}),
\end{eqnarray}
where $L_{cls}$ denotes the cross-entropy loss,  $L_{cen}$ denotes the center loss and $\lambda$ denotes the weight of $L_{cen}$. $Y^{i}$ is the predicted bilinear label vector based on features $F_{O}$, $F_{W}$ and $F_{E}$. $Y^{*}$ is the ground-truth label vector and $Y^{c}$ is the learned center vector of category.

\section{EXPERIMENTS AND DISCUSSIONS}

We evaluate the proposed approach on three publicly competitive fine-grained visual classification datasets, including CUB-$200$-$2011$~\cite{wah2011caltech}, Stanford Cars~\cite{krause20133d}, and FGVC Aircraft~\cite{maji13fine}. The detailed statistics with category numbers and the standard training/testing splits are summarized in Table~\ref{tab:datasets}. We employ top-1 accuracy as evaluation metric.

\begin{figure*}[ht]
\centering
\includegraphics[height=5cm]{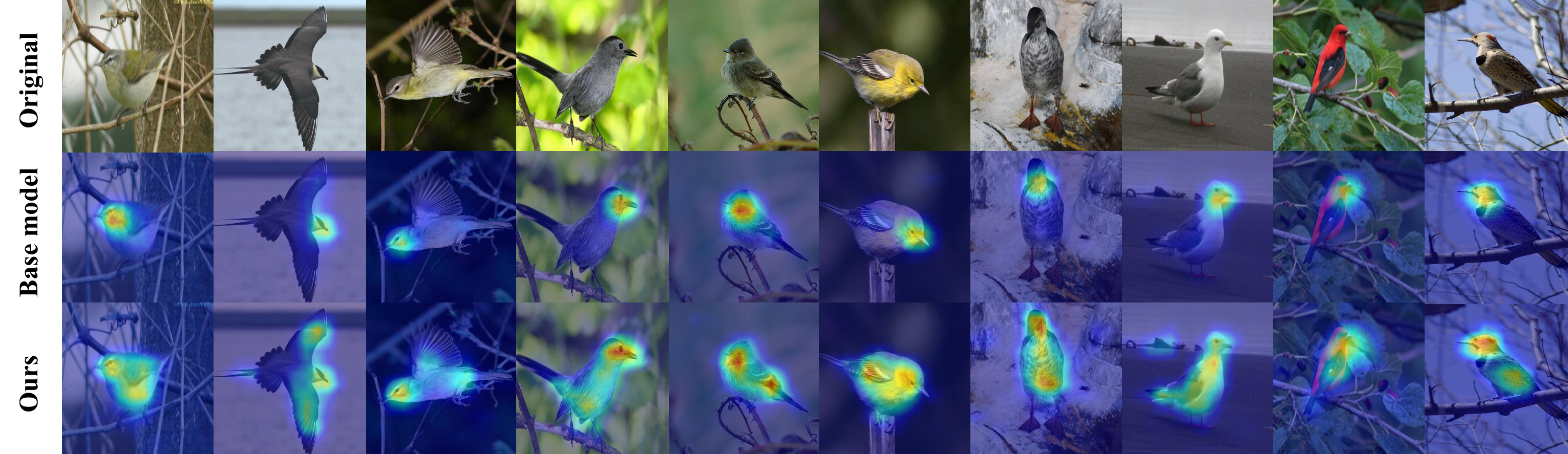}
\caption{The first line is the original images, the second line is the features learned by the base model, and the third line is the features learned by the proposed method. It can be observed that the proposed method has learned more feature information than the base model.}
\label{fig:visualization}
\end{figure*}

\subsection{Implementation details}
In all experiments, we use ResNet-$50$~\cite{he2016deep} and ResNet-$101$~\cite{he2016deep} as our base networks. We remove the last pooling layer and fully-connected layer, and add a billinear pooling~\cite{lin2015bilinear} to generate a more powerful feature representation. Then send the feature representation to a new fully-connected layer. The convolutional layers are pretrained on ImageNet~\cite{deng2009imagenet} and the fully connected layer is randomly initialized.

The input image size of all experiments is $448 \times 448$, which is the same as the most state-of-the-art fine-grained classification approaches~\cite{wang2018learning,yang2018learning,Zheng_2019_CVPR,sun2018multi,2020CIN}. In order to improve the generalization ability of the model, we implement data augmentation including random cropping and random horizontal flipping during training phase. Only center cropping is involved in inference phase. All experimental models are trained for $180$ epochs, using SGD as optimizer. The initial learning rate is set to $0.01$, which annealed by $0.9$ every $2$ epoches. We use a batch size of $32$, and the last $16$ samples belong to the same category as the first $16$ samples. The weight decay is set to $10^{-5}$ and the momentum is set to $0.9$.

\subsection{Experimental results}
The proposed method is compared with several representative methods without extra annotations on three fine-grained benchmark datasets of CUB-$200$-$2011$, Stanford Cars, and FGVC Aircraft, as shown in Table~\ref{tab:results}. The proposed method performs competitive results compared with the state of-the-art methods.

Specifically, the top-1 accuracy of the proposed method on CUB-$200$-$2011$ dataset is $88.9\%$, which is better than $88.6\%$ of API-Net~\cite{zhuang2020learning}. API-Net performs well because it also takes a pair of images as input, and learns a mutual feature vector by modeling the interaction between them to capture the semantic difference in the input pair. Our method is to capture the semantically similarity of a pair of same-category images, and disperse the network's attention to each discriminative area of the image.

On the other two datasets, our method also performs well, which achieves the result with $94.6\%$ top-1 accuracy for Stanford Cars dataset and $92.8\%$ top-1 accuracy for FGVC-Aircraft dataset. However, our method does not obtain the best results on these two datasets. Since cars and aircrafts are rigid objects, the intra-class variances of them are not assignificant as birds. And our method works better for objects that greatly vary within the category.
\begin{table}
\begin{center}
\caption{Ablation studies} \label{tab:Ablation studies}
\begin{tabular}{c|c|c|c}
  \hline
  CA-Module & AE-Module & center loss & accuracy(\%)  \\
\hline
\hline
  $\checkmark$ &  &  & $86.8$  \\
    % \hline
   & $\checkmark$ &  & $86.5$ \\
%   \hline
  $\checkmark$ &  & $\checkmark$ &  $87.9$ \\
%   \hline
    & $\checkmark$ & $\checkmark$ & $87.5$ \\
%   \hline
   $\checkmark$ & $\checkmark$ & $\checkmark$ & $88.3$  \\
  \hline
\end{tabular}
\end{center}
\vspace{-0.2cm}
\end{table}
\subsection{Ablation studies}
In order to verify the effectiveness of each component in the model, we conduct ablation studies on CUB-$200$-$2011$ dataset using ResNet-$50$ with some of components removed for better understanding the behavior of the model, as shown in Table~\ref{tab:Ablation studies}. 

It can be observed that CA-Module generates weighted feature maps through modeling the channel-wise interaction of same-category images, so that the network can learn their common features. AE-Module erases the most prominent area based on the weighted feature maps, making the informative areas learned by the network diversified. The center loss makes the extracted features more discriminative by constraining the distance between the feature vector and the center vector of the corresponding category. Each component of the model contributes to performance.
%\vspace{-0.3cm}
\subsection{Visualizations}
%\vspace{-0.2cm}
In order to further evaluate the effectiveness of our method, we apply Grad-CAM~\cite{Selvaraju_2017_ICCV} to visualize the images of the CUB-$200$-$2011$ dataset. Grad-CAM is formed by weighted summation of feature maps, which can show the importance of each area to its classification. We compare the visualization results of our method with the base model (ResNet-$50$), as shown in Figure~\ref{fig:visualization}. It can be observed that the base model only learns the most prominent area of the image, such as the bird's beak. Our method can learn more abundant and discriminative features, including wings and claws. This is because that our method can distribute attention to each area to make the prediction more comprehensive, which can not only focus on the salient features, but also capture the subtle and fine-grained features.

\section{CONCLUSIONS}
We propose an effective fine-grained visual classification method, namely progressive co-attention network. Among them, the co-attention module learns discriminative features by comparing same-category images, and the attention erase module learns subtle complementary features of images by erasing the most prominent area. We have conducted experiments on CUB-$200$-$2011$, Stanford Cars, and FGVC Aircraft datasets, which are superior to most existing methods. 

\vspace*{0.1in}
\noindent{\textbf{Acknowledgments}: This work was supported in part by the National Key R \& D Program of China under Grant $2019$YFF$0303300$ and under Subject II No.$2019$YFF$0303302$, in part by National Natural Science Foundation of China (NSFC) No.$61922015$,$61773071$, U$19$B$2036$, in part by Beijing Natural ScienceFoundation Project No. Z$200002$.}

\bibliographystyle{IEEEbib}
\bibliography{VCIP2021}

\end{document}